\documentclass[conference]{IEEEtran}
\IEEEoverridecommandlockouts

\usepackage{cite}
\usepackage{amsmath,amssymb,amsfonts}
\usepackage{algorithmic}
\usepackage{graphicx}
\usepackage{textcomp}
\usepackage{xcolor}
\usepackage{hyperref}
\usepackage{booktabs}
\usepackage{tabularx}
\usepackage{array}
\usepackage{tagging}

\usepackage{sc26repro}

\def\BibTeX{{\rm B\kern-.05em{\sc i\kern-.025em b}\kern-.08em
    T\kern-.1667em\lower.7ex\hbox{E}\kern-.125emX}}

\begin{document}

\title{Multi-Agent Collaboration for Automated Design Exploration on High Performance Computing}

\author{%
\IEEEauthorblockN{%
Harshitha Menon\IEEEauthorrefmark{1},
Charles Jekel\IEEEauthorrefmark{1},
Kevin Korner\IEEEauthorrefmark{1},
Giselle Fernandez\IEEEauthorrefmark{1},
Brian Gunnarson\IEEEauthorrefmark{1},
Nathan Brown\IEEEauthorrefmark{2},\\
Michael Stees\IEEEauthorrefmark{1},
Walter Nissen\IEEEauthorrefmark{1},
Meir Shachar\IEEEauthorrefmark{1},
Dane Sterbentz\IEEEauthorrefmark{1},
William Schill\IEEEauthorrefmark{1},
Yue Hao\IEEEauthorrefmark{1},
Robert Rieben\IEEEauthorrefmark{1},\\
William Quadros\IEEEauthorrefmark{2},
Steve Owen\IEEEauthorrefmark{2},
Scott Mitchell\IEEEauthorrefmark{2},
Ismael Boureima\IEEEauthorrefmark{3},
Jonathan Belof\IEEEauthorrefmark{1}\IEEEauthorrefmark{4}%
}
\IEEEauthorblockA{%
\IEEEauthorrefmark{1}Lawrence Livermore National Laboratory, Livermore, CA, USA\\
\IEEEauthorrefmark{2}Sandia National Laboratories, Albuquerque, NM, USA\\
\IEEEauthorrefmark{3}Los Alamos National Laboratory, Los Alamos, NM, USA\quad
\IEEEauthorrefmark{4}AMD\\
Correspondence: \texttt{harshitha@llnl.gov}%
}
}

\maketitle

\begin{abstract}
Scientific discovery requires rapidly testing hypotheses, generating results,
and learning from them, yet most of it is
heavily manual. This is because, researchers must coordinate simulation and analysis
through manually written scripts and explicit workflows. We
present MADA (Multi-Agent Design Assistant), a multi-agent framework for scientific design exploration
with expert-in-the-loop interaction. Scientists describe their objectives in natural language, and MADA
dynamically orchestrates the design loop through specialized agents. We
showcase MADA interfacing with production tools including Flux, Cubit, PMesh, Laghos, and MARBL
on a leadership-class HPC system. 
We evaluate on three hydrodynamic instability problems: iterative
RMI design with Laghos, surrogate-based exploration for high-velocity
impact, and an 1,800-run adaptive RTI campaign with MARBL. In our evaluations,
agents suggest design parameters to explore, launch jobs, analyze intermediate results, and adapt exploration
strategies.
We share observations and operational lessons
from this deployment to help practitioners build Agent-driven scientific
workflows on HPC systems.

\end{abstract}

% \begin{IEEEkeywords}
% multi-agent systems, large language models, high performance computing, scientific workflows, design optimization
% \end{IEEEkeywords}

\section{Introduction}
\label{sec:intro}
The scientific discovery cycle increasingly depends on complex workflows that
integrate simulation, analysis, and decision-making. From fusion energy research
to novel materials discovery, progress requires rapidly testing hypotheses,
generating results, and learning from them. Although high-performance computing
(HPC) has made it possible to simulate physics with high fidelity, but simply
running larger simulations is not sufficient. What is needed are closed-loop 
workflows that tightly integrate simulation, analysis, and design exploration.

Today, however, most design workflows are cumbersome and heavily manual. Even
a single design study under strict physical and engineering constraints
requires moving through an expensive cycle of design, simulation, validation,
and analysis. Although workflow management systems~\cite{deelman2015pegasus,wilde2011swift,jain2015fireworks,PETERSON2022255}
automate job execution and data movement, they still place a significant manual
burden on scientists. In particular, they must explicitly construct workflow graphs,
define how simulations are executed, and specify exploration strategies.
While these systems support job scheduling, they lack the complex orchestration needed for
mesh generation, adaptive sampling, and iterative refinement based on intermediate results.
As a result, researchers spend much of their effort managing orchestration
logic rather than advancing the underlying scientific objectives.
Therefore, we need automated and intelligent design workflows that can
reason about results, adapt strategies, and coordinate diverse tools to
accelerate the pace of discovery.

Large Language Models (LLMs) have shown impressive capability in several
domains, from natural language understanding and code generation to planning and
problem solving~\cite{brown2020language,achiam2023gpt,wei2022chain}. 
Their ability to reason over complex information makes them
attractive to scientific applications~\cite{boiko2023autonomous,m2024augmenting,taylor2022galactica,wang2023scibench}. 
However, the best way to leverage LLMs
for complex scientific and engineering tasks remains an open question.

This paper presents MADA (Multi-Agent
Design Assistant), a multi-agent framework powered by LLMs and integrated with
domain-specific tools. MADA unifies LLM-based reasoning, physics simulation,
HPC job orchestration, and scientific tool integration in a single system for
iterative scientific workflows. Unlike traditional workflow systems that require
explicit workflow graphs and hand-written scripts, MADA accepts natural language
problem descriptions and dynamically orchestrates the design loop.
The framework modularizes the scientific design workflow into
specialized agents, each responsible for different stages of the scientific workflow.
Within this framework, LLMs act as reasoning engines that guide collaboration by analyzing results,
extracting insights, and adapting workflows as the context evolves.
MADA also supports expert-in-the-loop interaction through its natural
language interface. Experts can specify objectives and constraints, review
intermediate results, and evaluate proposed designs.

We evaluate MADA on hydrodynamic instability problems critical to Inertial Confinement
Fusion (ICF) research: Richtmyer--Meshkov Instability (RMI)~\cite{schill2024suppression}
and Rayleigh--Taylor Instability (RTI). Both instabilities can degrade capsule integrity
and prevent successful fusion ignition. Understanding them requires exploring how interface
geometry, material properties, and drive conditions interact, a process that requires tight
coordination of mesh generation, simulation, and analysis. The framework consists of several 
agents orchestrated by a planning component that coordinates cycles of design, simulation, 
and analysis. Specifically, the Job Management Agent (JMA) handles large ensembles of jobs 
on HPC systems via Flux~\cite{ahn2020flux}, the Geometry Agent (GA) automates mesh generation 
through Cubit~\cite{blacker2016cubit} or PMesh~\cite{hardin1994pmesh}, and the Inverse Design
Agent (IDA) explores the design space, guided by simulation results. Together,
these agents transform what has traditionally been a cumbersome, manual workflow
into an automated, modular, and scalable approach to scientific design exploration.
Across our evaluations, MADA drives iterative design with
HPC simulations, performs surrogate-based exploration, and steers a
1800-run adaptive campaign on Tuolumne, a leadership-class HPC system.

The primary contributions of this paper are as follows:
\begin{itemize}
    \item MADA, a multi-agent framework that decomposes scientific design workflows
    into specialized agents for mesh generation, job orchestration, and inverse design
    on HPC systems.
    \item Integration with existing HPC tools and infrastructure via the Model Context
    Protocol, which allows agents to directly call domain-specific tools.
    \item An evaluation on ICF-relevant instabilities across three settings:
    (a) iterative RMI design with Laghos~\cite{dobrev2012} on HPC,
    (b) surrogate-based exploration with a pre-trained ML model~\cite{jekel2024machine}, and
    (c) adaptive RTI campaign steering with MARBL~\cite{rieben2020marbl}, where MADA
    autonomously managed a 1800-run campaign.
\end{itemize}

As AI capabilities continue to advance, MADA demonstrates how LLM agents can be effectively
integrated with HPC tools and infrastructure to support scientific workflows. We present the
framework, along with observations and operational insights,
to inform scientific workflow developers and HPC practitioners seeking to implement agentic workflows.

\section{Background}
\label{sec:background}
LLMs have demonstrated remarkable capabilities in
understanding and generating human language, code, and structured
reasoning~\cite{brown2020language,achiam2023gpt,touvron2023llama}. As a result, recent
advances have shown that LLMs can serve as powerful reasoning engines for
complex tasks that require planning, tool use, and multi-step problem-solving~\cite{yao2023react,shinn2023reflexion,wei2022chain}. In scientific
computing contexts, LLMs provide a natural interface between human intent and
computational workflows, translating high-level objectives into executable
actions~\cite{boiko2023autonomous}.

\subsection{Agents and Tool Use}

An LLM-based agent perceives its environment, makes decisions, and takes
actions to achieve specific goals. It extends LLMs
with tool use, memory, and goal-directed
behavior~\cite{schick2023toolformer,qin2023toolllm,patil2024gorilla}, which 
turns them into active problem solvers that can reason and iterate on
solutions~\cite{wang2023voyager,park2023generative,shinn2023reflexion,madaan2023self,yao2023tree,hao2023reasoning}. 
A key characteristic relevant to HPC is tool integration, where agents can call 
specialized tools for tasks beyond the LLM's native capabilities, 
including running simulations, generating meshes, or analyzing data~\cite{ruan2023tptu,yang2023intercode}.
Moreover, their ability to
understand context, maintain state across interactions, and adapt to feedback
makes them well-suited for orchestrating scientific workflows that
traditionally required extensive manual
oversight.

\subsection{Multi-Agent Systems}

While single agents can be effective for narrowly scoped tasks, complex scientific workflows often
require diverse expertise and parallel execution capabilities that exceed what
a single agent can efficiently manage~\cite{hong2023metagpt,chen2023agentverse}.
Multi-agent systems address this by orchestrating multiple specialized agents
that collaborate to solve larger
problems~\cite{wu2024autogen,li2023camel,du2023improving}. Each agent focuses on
its area of expertise while contributing to the collective
goal~\cite{qian2023communicative,zhang2023proagent}. Multi-agent architectures
offer several advantages. First, specialization allows each agent to be optimized for
specific tasks, improving overall system performance and
reliability~\cite{hong2023metagpt}. Second, modularity makes it possible to add
new agents or modify existing ones without disrupting the whole system~\cite{wu2024autogen}.
Third, distributing responsibilities across multiple agents helps them focus on
relevant context and manage the limited context windows of LLMs.
Finally, this division of labor improves robustness,
as failures in one agent do not compromise the entire workflow~\cite{zhang2023building}.

\subsection{Model Context Protocol (MCP)}

MCP is an open standard that enables integration
between LLMs and external tools. MCP follows a
client-server architecture where servers
expose Tools (executable functions), Resources (contextual data), and 
Prompts (reusable templates). Communication uses JSON-RPC 2.0 over 
standard input/output or Streamable HTTP.

Within our framework, MCP serves as the communication backbone between agents
and tools. Each MCP server wraps a specific tool (Flux, Laghos, MARBL, Cubit,
surrogate models) and provides a standardized interface that agents can query.
This separation means MCP servers can be reused with different agent
frameworks without modification.

\section{MADA System Overview}
\label{sec:system}
Modern scientific discovery requires solving problems that a single tool or
agent cannot address alone. While a single agent can effectively
handle narrow, well-bounded tasks such as answering queries, retrieving data, or
performing basic calculations, it quickly becomes overloaded when faced with
complex, multi-step workflows. Scientific workflows is one such scenario, demanding diverse skill sets,
long-horizon reasoning, and coordination across various environments such as 
simulation code, domain-specific tools, and HPC systems.

\begin{figure}[ht]
    \centering
    \includegraphics[width=\columnwidth]{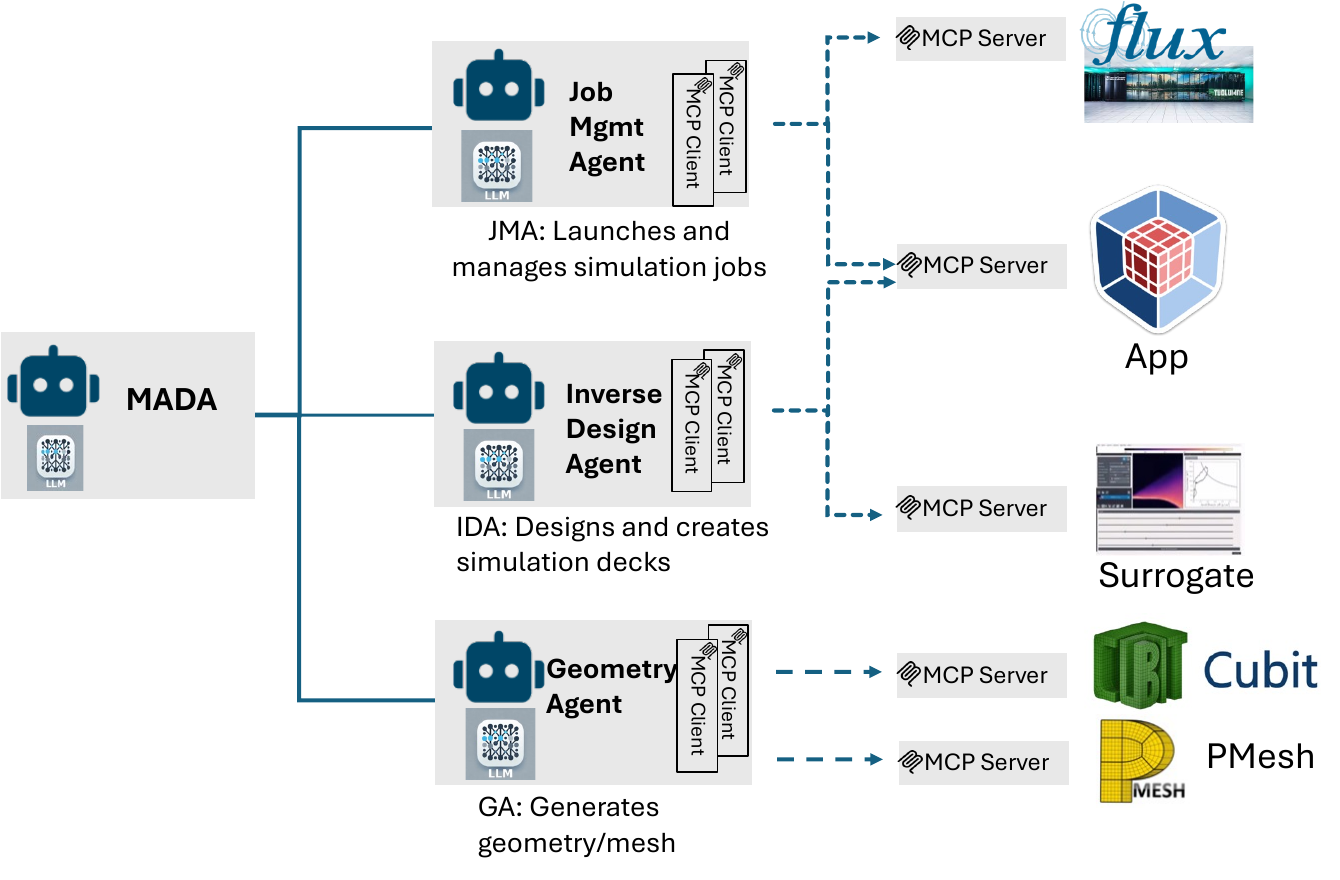}
    \caption{%
        MADA system architecture. The framework orchestrates three specialized
        agents: the JMA manages simulations on HPC via
        Flux, the GA generates meshes through Cubit or PMesh, and the
        IDA explores the design space. Agents communicate
        with tools via MCP.}
    \label{fig:system_overview}
\end{figure}

The main goal of MADA is to automate iterative scientific design loop, which
includes proposing candidate designs, generating meshes, running simulations,
analyzing results, and refining. To support this process, MADA uses a set of specialized agents, each responsible
for a distinct role in the workflow (Figure~\ref{fig:system_overview}). The JMA handles simulation
execution on HPC systems, the GA produces meshes, and the IDA analyzes results and proposes new candidates. These agents access external tools through
MCP (Section~\ref{sec:background}), which provides a
uniform interface for tool discovery and invocation. 

A typical design exploration proceeds as follows. The user specifies the
objective (e.g., suppress RMI), design variables (e.g., interface geometry
parameters), and constraints (e.g., parameter bounds). Next, the GA generates
the corresponding problem mesh. The IDA then produces an initial batch of candidate designs, either through
broad exploration or targeted sampling near promising regions. For each
candidate, the JMA submits the simulation. Once the results return, IDA extracts quantities of interest,
ranks configurations, and proposes the next batch. This cycle repeats until the
user terminates or a convergence criterion is met.

More broadly, we build our framework around five core components:
\begin{itemize}
    \item Specialized Agents: Agents responsible for different stages of the
    scientific workflow.
    \item Planning: A planning component decomposes high-level scientific goals into
    subtasks.
    \item Coordination: Agents communicate through well-defined interfaces to
    schedule work and share results.
    \item Tool Integration: MCP expose
    specialized scientific tools (mesh generators, simulation codes, HPC
    schedulers) through a standardized interface for agent interaction.
    \item Memory: Results and context persist to make possible continuous
    improvement over iterative design cycles.
\end{itemize}

MADA's HPC integration spans the complete simulation pipeline using
production-grade tools: Flux for exascale job scheduling, Cubit and
PMesh for mesh generation, Laghos (an Exascale Computing Project mini-app),
and MARBL (a multi-physics production code). Managing meshing, ensembles of
simulations on leadership-class systems, including job
submission, monitoring, failure recovery, and adaptive continuation, is
precisely what makes HPC workflows challenging. MADA automates this
end-to-end orchestration.

\subsection{Job Management Agent (JMA)}
\label{sec:jma}
The JMA interfaces with simulation codes and ensures
reliable execution of large simulation ensembles on HPC systems. It abstracts
the complexity of interacting with schedulers like Flux and provides a robust
interface to launch, monitor, and collect simulation results.
The JMA utilizes two key types of MCP servers: simulation codes and schedulers.
The simulation code MCP servers define what to run, while the scheduler MCP
servers define how to execute the jobs that simulation codes specify.
%  (see Figure~\ref{fig:jma_architecture}).
% \begin{figure}[t]
%     \centering
%     \includegraphics[width=\linewidth]{figs/jma-architecture.png}
%     \caption{%
%         The architecture of the JMA, showcasing both the simulation codes MCP servers
%         and the scheduler MCP servers.
%     }
%     \label{fig:jma_architecture}
% \end{figure}

% We designed the JMA so that users can easily integrate additional job scheduler
% MCP servers (for example, Slurm, Merlin, or Parsl) and other simulation codes. A
% JSON configuration file configures the JMA, GA, and IDA by pointing each agent
% to the appropriate MCP servers. For the JMA, this currently means Laghos for
% simulation and Flux for job scheduling. To support a new scheduler or code,
% users add a corresponding MCP server entry to this configuration
% file if one already exists, or wrap the tool in a new MCP server.

\subsubsection{Simulation Code Interface}

\paragraph{Laghos}
\label{sec:laghos}
Laghos (LAGrangian High-Order Solver)~\cite{ceed_laghos_2017} is an open-source miniapp code
% derived from MARBL~\cite{rieben2020marbl} 
that solves the time-dependent Euler equations of compressible gas dynamics in a moving Lagrangian frame 
using unstructured high-order finite-element spatial discretization and explicit high-order time stepping. 

In this work, we employ Laghos3~\cite{korner2026differentiable}, the differentiable extension of Laghos with the following extensions:

\begin{itemize}
  \item Support for complex material models, such as Mie-Gruneisen equations of state
  \item A unified interface for specifying initial conditions, boundary conditions, and geometries
  \item Configuration of problem setups through Lua-based input decks
\end{itemize}

These enhancements enable fine-grained, programmatic control of problem configuration and parameters via the JMA. 
The Lua-based configuration is particularly useful for agentic workflows: the JMA can pass new parameters 
at runtime without needing to recompile the application for each exploration.

We expose Laghos functionality through an MCP server that provides tools for
staging and configuring simulations. The primary tool, \texttt{generate\_runs()},
takes a set of design parameters and generates the corresponding simulation
input files, including Lua configuration decks and stages them for execution. This tool
returns a list of run descriptions that the JMA can pass to the Flux scheduler
for execution. In this design, the JMA delegates application configuration and staging
to application-specific MCP server and then forwards the resulting run
descriptions to a scheduler-specific MCP server, thereby maintaining a clear
separation of concerns.

\paragraph{MARBL}
\label{sec:marbl}
MARBL~\cite{rieben2020marbl} is a production multi-physics simulation code
that supports arbitrary Lagrangian-Eulerian (ALE) hydrodynamics with
high-order finite elements on unstructured meshes and runs on GPU-accelerated
HPC systems.

We expose MARBL through an MCP server that provides tools for
parameter-sweep construction, QoI analysis, and plotting. The tool
\texttt{prepare\_parameter\_sweep} generates simulation input directories and
configuration files from a set of continuous and discrete design parameters. As with
Laghos, the JMA delegates simulation setup to the MARBL MCP server and
forwards the resulting run descriptions to the Flux scheduler for execution.

\subsubsection{Scheduler Integration}

We architected the JMA for modular scheduler integration. 
The JMA abstracts scheduler-specific logic behind a unified interface, allowing
users to switch between job schedulers and extend to new ones without disrupting
existing workflows. This modular design supports scalable simulation workflows
across diverse HPC environments.

Currently, the JMA supports the Flux job scheduler for managing jobs. Flux is a
next-generation, hierarchical resource management and job scheduling framework
developed for HPC systems~\cite{ahn2020flux}. It provides a flexible and composable scheduling architecture
that supports dynamic partitioning and allocation of resources across multiple nested
scheduling domains. By treating scheduling as a distributed service, Flux enables
fine-grained control, rapid job dispatch, and efficient resource utilization on
systems ranging from clusters to exascale supercomputers. 
% Its modular design,
% rich APIs, and integration with modern computing environments make it a robust
% foundation for advanced workload management and experimentation.

Flux integration is implemented as an MCP server that exposes a set of
MCP tools built using the Python bindings of Flux. The server provides three key MCP tools: \texttt{submit\_job()},
\texttt{submit\_jobs\_async()}, and \texttt{check\_job\_status()}.
% , and \texttt{execute\_generated\_runs()}. 
The \texttt{submit\_job()} tool submits a single
command to the Flux scheduler, allowing users to specify basic resource requirements
such as the number of nodes, tasks per node, time limit, job name, and working directory,
and returns the corresponding Flux job ID. The \texttt{submit\_jobs\_async()} tool accepts
a JSON description of multiple runs and submits them asynchronously, immediately returning
job identifiers so that large ensembles or parameter sweeps can be launched without blocking.
The \texttt{check\_job\_status()} tool queries the scheduler for the state of a given job, or
all managed jobs if the user provides no ID, and returns a JSON summary of their current status.
% Finally, \texttt{execute\_generated\_runs()} takes generated run descriptions (for example,
% from the Laghos MCP server) and executes them synchronously, submitting and monitoring each
% run until completion and returning a textual summary of their outcomes.

\subsection{Geometry Agent (GA)}

\label{sec:ga}
The GA automates geometry generation and mesh preparation for
candidate designs by translating plain-text CAD descriptions into commands for one of two mesh generation packages: Cubit,
and PMesh.

Cubit is Sandia National Laboratories' flagship geometry-creation and mesh-generation tool. It is used to build solid models
(points, lines, surfaces, volumes) and to produce high-quality finite-element meshes. 
Cubit exposes Python API, which is a thin wrapper around its domain-specific language (DSL).
GA takes natural-language prompts, drives Cubit through this API, and can create geometry, 
apply mesh controls, and retrieve mesh-quality metrics.

PMesh is Lawrence Livermore's massively parallel CAD-based mesher for multi-physics HPC simulation. 
While CUBIT has extensive CAD creation and unstructured mesh generation, PMesh focuses on 
precise control over mesh structure and resolution while generating meshes of billions of elements. 
All scripting is done through its Python API, which is similar to Cubit's. 
The output is an MFEM-formatted~\cite{andrej2024mfem} cubic NURBS mesh, suitable for simulation as-is, 
or able to be refined in parallel with no loss of accuracy.

\subsubsection{Leveraging Documentation via RAG}
The GA relies on a Retrieval-Augmented Generation (RAG) pipeline that draws from a curated knowledge base
containing the Cubit and PMesh user-documentation and a collection of vetted example scripts.
For Cubit, each user-manual section was pre-processed into discrete text chunks, each encapsulating a one--three DSL function names, concise descriptions
of purpose and parameters, and representative usage examples.
For PMesh, documentation is programmatically generated via introspection for 1500 function points. Each of them contain a high-level description, names and types of positional and keyword arguments, extended prose description, and one or more examples.
GA embeds the request using a vector-encoding model and then computes 
cosine similarity between this query vector and the vectors of all documentation chunks. 
For PMesh, we use a hybrid search by augmenting vector search with Bm25 keyword matching~\cite{robertson2009bm25}.
The top-N most relevant chunks are given to the LLM as context. By presenting the model with the most relevant function definitions and examples, this step guides generation toward syntactically correct and semantically appropriate DSL.

\subsubsection{Capturing Spatial and Geometric Information}

A critical aspect of any CAD modeling task is the ability to understand the geometric and spatial attributes of the present model.
Achieving such an understanding with an LLM is a non-trivial task. Previous works have shown vision language models are able to provide
limited understanding of a CAD model, but commonly suffer from significant information loss when the CAD views are not sufficiently detailed~\cite{khan2024leveraging, alrashedy2024generating}.
Therefore, we leverage a topological approach to feed general geometric and spatial information into the GA. We construct a geometric graph that 
encodes the complete topology of the active Cubit model.  Each node in the graph corresponds to a fundamental entity (i.e. vertex, edge, surface)
and is enriched with attribute fields that capture its quantitative description. For a vertex node we store its Cartesian coordinates,
for an edge node we record its length, centroid coordinates, and mesh size/intervals, and finally surface nodes define centroid coordinates, area, outward normal vector and mesh size/intervals. 
The graph edges represent the incidence relationships: a vertex node is linked to every edge node that uses it, an edge node connects to the surface nodes that contain it. This explicit connectivity preserves the full topological hierarchy without requiring additional inference.

After the graph is assembled, we serialize it into a structured textual representation that lists each entity together 
with its attributes and adjacency lists.  The format follows a consistent schema, ensuring that the language model can parse the data deterministically.
This structured block is then inserted into the system prompt of the GA, providing the LLM with an up-to-date snapshot of the model's geometry and topology.
With this information readily available, the GA can generate precise Cubit DSL commands, resolve spatial queries (such as ``find the nearest surface to vertex 5''), and make informed decisions about geometry modifications or mesh refinements.

\subsubsection{Specialized Design Tools}

Beyond general-purpose mesh generation, the GA can invoke specialized tools
exposed through the Cubit and PMesh MCP servers. For example, the
\texttt{generate\_rmi\_mesh} tool parametrically designs and exports meshes
for RMI studies, accepting interface perturbation parameters and producing
simulation-ready meshes. For defeaturing operations, the Cubit MCP server
exposes power tools that remove small cavities or fillets according to
user-specified tolerances. The tool-calling framework is extensible: new
capabilities can be registered by adding tools to the corresponding MCP
server. The GA queries available tools at runtime and falls back to
alternative methods if a required tool is absent.

\subsubsection{Verification and Iteration}
For Cubit, we use a simple axis-aligned bounding box to
assess approximate geometric congruence with a human-generated reference
result. We also compare the textual similarity of the response to
a human-generated reference command script. To test a PMesh
result we generate a best-match bijection that uses topological
correspondence (vertex count, connectivity) as well as approximate geometric
congruence (maximum distance). In particular, we do not penalize
the GA for producing additional geometry, or proceeding further
than requested, such as mesh generation.

When tool calls or generated commands fail to produce the
reference result, which is inevitable, clear and specific error
messages are essential. In cases where the initially provided
context was insufficient, reporting a simple syntax error can
prompt the GA to retrieve new, hopefully more relevant
context. Similarly, reporting that the geometry deviated from the
desired result may allow the GA to go back and attempt to correct
its error.

\subsection{Inverse Design Agent (IDA)}
\label{sec:ida}
% The goal of the Inverse Design Agent (IDA) is to explore the design space and
% identify configurations that optimize a specified objective. Given simulation
% results or surrogate model predictions, the IDA extracts quantities of interest
% (QoI), ranks candidate designs, and proposes new configurations for evaluation.

% For RMI suppression, the primary QoI is jet length---the extent of material
% jetting at the perturbed interface. When working with full hydrodynamics
% simulations, the IDA extracts jet length from tracer diagnostics in the
% simulation output. When using surrogate models, it computes jet length from
% predicted density fields. The IDA ranks candidates by their QoI values and uses
% LLM-based reasoning to identify patterns and propose promising new
% configurations.

% The IDA interfaces with machine learning surrogate models for rapid design
% exploration~\cite{jekel2024machine}. These surrogates approximate the simulation
% response at a fraction of the computational cost. We expose both QoI calculation
% and surrogate model inference as MCP tools that the IDA invokes through a
% standardized interface. This architecture allows the IDA to use full-fidelity
% HPC simulations when accuracy is critical and fast surrogate evaluations when
% exploring large design spaces.

The primary task of IDA is optimization and design-space
exploration. It analyzes simulation results and, when available, uses surrogate 
models to approximate simulation response at a fraction of the computational cost. 
The IDA calculates quantities of interest (QoI) 
that is used to guide design decisions. For RMI suppression problems, the QoI is jet length,
which measures the extent of material jetting at the perturbed interface. When
working with full hydrodynamics simulations, the IDA extracts jet length from
tracer diagnostics in the simulation output. When using surrogate models, it
computes jet length from the predicted density fields. The IDA ranks candidate
designs by their QoI values and identifies configurations that best meet the
design objectives.

% The IDA interfaces with machine learning surrogate models for rapid design
% exploration. These surrogate models approximate the simulation
% response at a fraction of the computational cost. The IDA queries the surrogate at random points of
% the parameter space, computes QoI from predicted fields, and analyzes results using LLM-based
% reasoning to propose promising new configurations.

The IDA connects to two types of MCP servers depending on the evaluation mode.
For simulations, the IDA connects to an application-specific MCP
server that exposes tools to calculate QoI from simulation outputs. For example,
the Laghos MCP server provides \texttt{get\_qoi()}, which extracts jet length from tracer
diagnostics, and the MARBL MCP server provides \texttt{rank\_rti\_mixing\_study},
which computes and ranks the mixing QoI across a completed study.
For surrogate-based exploration, the IDA connects to a surrogate
model~\cite{jekel2024machine} MCP server that exposes \texttt{get\_objective()}, which runs inference and computes objective
values for a given design input. This separation allows us to swap evaluation
backends without modifying the IDA logic, and lets the IDA use HPC simulations
when accuracy is critical or fast surrogate models to efficiently explore large
design spaces.

\subsection{System Orchestration}
\label{sec:orch}
MADA orchestrates scientific workflows by converting high-level user
objectives into executable plans and coordinating them across specialized agents.
Unlike traditional workflow management systems that require explicit specification
of every step, MADA uses LLM-based reasoning to dynamically decompose tasks,
allocate work to appropriate agents, and adapt the execution strategy based on
intermediate results.

We implement MADA on top of the Microsoft Agent Framework (formerly
AutoGen)~\cite{wu2024autogen}, a platform for building multi-agent applications.
A JSON configuration file defines each agent's role, description, and the MCP
servers it connects to, making it straightforward to add new agents or
reconfigure existing ones for different studies.
% Figure~\ref{fig:system_coordination} illustrates the coordination process, which
% operates in three stages. 
The coordination process operates in three stages.
First, a context analyzer consolidates the conversation
history and agent capabilities. Next, a selector uses this context to determine
which agent should act next, adapting to the current workflow state.
%  rather than following a fixed sequence. 
Finally, the selected agent produces a response, often
invoking one or more MCP tools, and shares a summarized result with all agents.
As a result, the entire system maintains a consistent view of the workflow state.
% When an agent fails or encounters unexpected behavior, the system adapts by
% adjusting subsequent actions. 
This multi-agent coordination improves modularity, but introduces
trade-offs: each handoff between agents adds latency and the potential for
miscommunication.% (e.g., parameter format mismatches between the IDA and JMA),
% and an incorrect selector decision can waste an LLM call by invoking the wrong
% agent for the current workflow state.

% \begin{figure}[t]
%     \centering
%     \includegraphics[width=0.8\linewidth]{figs/system_coordination.pdf}
%     \caption{%
%         Coordination process. The context analyzer processes conversation
%         history and agent roles. The selector chooses the next speaker. The
%         selected agent responds, and the system broadcasts the message to maintain
%         shared state. This cycle repeats until the system reaches termination.}
%     \label{fig:system_coordination}
% \end{figure}

Agents interact with external tools through MCP servers that expose standardized
interfaces. Each tool (Flux, Cubit, PMesh, simulation codes, surrogate models) is
wrapped in an MCP server. Agents discover available capabilities and invoke
functions by sending structured requests. MCP handles parameter validation, error
reporting, and result formatting. This separation between agent logic and tool
implementation means that MCP-compliant tools interoperate with any MCP-compliant
agentic framework. Table~\ref{tab:mcp_interfaces} summarizes the servers and
functions available for our design workflows.

\begin{table}[t]
\caption{MCP servers and the functions they expose to agents.}
\label{tab:mcp_interfaces}
\centering
\scriptsize
\begin{tabularx}{\linewidth}{@{}>{\raggedright\arraybackslash}p{0.11\linewidth}>{\raggedright\arraybackslash}p{0.45\linewidth}X@{}}
\toprule
Server & Function & Description \\
\midrule
Flux & \texttt{submit\_job} & Single job submission \\
Flux & \texttt{submit\_jobs\_async} & Async ensemble submission \\
Flux & \texttt{check\_job\_status} & Query job state \\
\midrule
Laghos & \texttt{generate\_runs} & Stage simulation inputs \\
Laghos & \texttt{get\_qoi} & Extract jet length \\
\midrule
MARBL & \texttt{prepare\_parameter\_sweep} & Generate sweep dirs \& inputs \\
MARBL & \texttt{rank\_rti\_mixing\_study} & Compute \& rank $Q_{\mathrm{mix}}$ \\
MARBL & \texttt{plot\_rti\_density\_snapshots} & Density snapshot panels \\
MARBL & \texttt{plot\_rti\_parameter\_comparison} & Parameter comparison plots \\
\midrule
Surrogate & \texttt{get\_objective} & ML inference \\
\midrule
Cubit & \texttt{generate\_rmi\_mesh} & Parametric RMI mesh design \& export \\
Cubit & Power tools & Defeaturing operations \\
Cubit & Python API & Mesh generation \\
\midrule
PMesh & \texttt{generate\_rmi\_mesh} & Parametric RMI mesh design \& export \\
PMesh & Python API & Parallel mesh generation \\
\bottomrule
\end{tabularx}
\end{table}

\subsection{Expert-in-the-Loop with MADA}
\label{sec:expert}
Although MADA can operate autonomously, the domain experts can play a role in
guiding the design exploration process. MADA supports expert-in-the-loop
interaction via two interfaces. A command-line interface (CLI)
lets experts interact with MADA directly from a terminal, suitable for
HPC environments where graphical interfaces may not work. We also
provide a graphical user interface (GUI) that displays simulation results,
optimization progress, and agent reasoning. Through both the
CLI and GUI, experts can interact at multiple stages of the workflow.
At the start of the exploration,
experts specify design objectives, constraints, and variable ranges.
During execution, they may review intermediate results, adjust
search parameters, and redirect the exploration based on their domain knowledge.
At the end of each iteration, experts can evaluate the designs
that MADA proposes and decide whether to continue refining, explore other strategies, or terminate the study.
This interaction model combines the strengths of automated exploration with
human expertise.

\section{Evaluation and Case Studies}
\label{sec:eval}
ICF is an approach to fusion energy in which a
small fuel capsule is compressed by laser until fusion ignition
occurs. A key challenge is that hydrodynamic instabilities at material interfaces
can disrupt the implosion and prevent ignition~\cite{desjardins2019platform}.
We evaluate MADA on two such instabilities: RMI, which occurs when a shock wave amplifies
perturbations at a material interface, causing large jet-like
growths~\cite{chen2019effects,sternberger2017comparative,buttler2012unstable},
and RTI, which develops when a heavy fluid
accelerates into a lighter one. Understanding how design parameters affect
instability growth requires exploring how interface geometry, material
properties, and other conditions interact, exactly the kind of problem MADA
is designed to address.

We ran all evaluations on Tuolumne, a leadership-class system at Lawrence Livermore National
Laboratory containing 1,152 nodes, each with 4th Generation AMD EPYC processors
(96 cores), four AMD MI300A accelerators, and 512~GB of memory. The system
delivers a peak performance of 294 petaflops and debuted at \#10 on the November
2024 Top500 list. Tuolumne uses Flux~\cite{ahn2020flux} as its workload manager,
a hierarchical resource manager designed for exascale workflows.

\subsection{Design Exploration with Laghos}
\begin{figure}[t]
    \centering
    \includegraphics[width=0.95\linewidth]{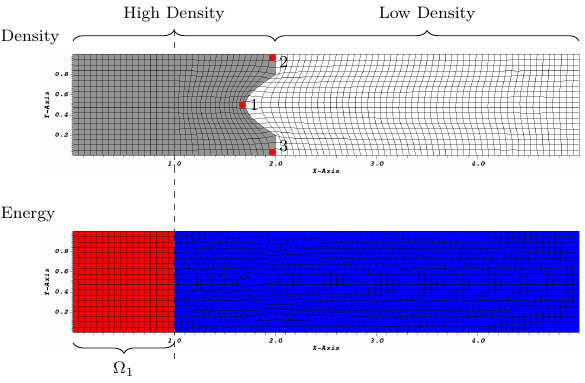}
    \caption{%
        Simulation setup for 2D RMI~\cite{korner2026differentiable}. Top: mesh with
        perturbed interface generated by the GA. Bottom: energy initialization
        defined by the sampled design parameters.
    }
    \label{fig:2d_rmi_laghos_setup}
\end{figure}

In this evaluation, we exercise the full design loop: specify design variables, generate meshes,
run simulations, analyze results, and propose refinements.

\begin{figure}[t]
    \centering
    \includegraphics[width=\linewidth]{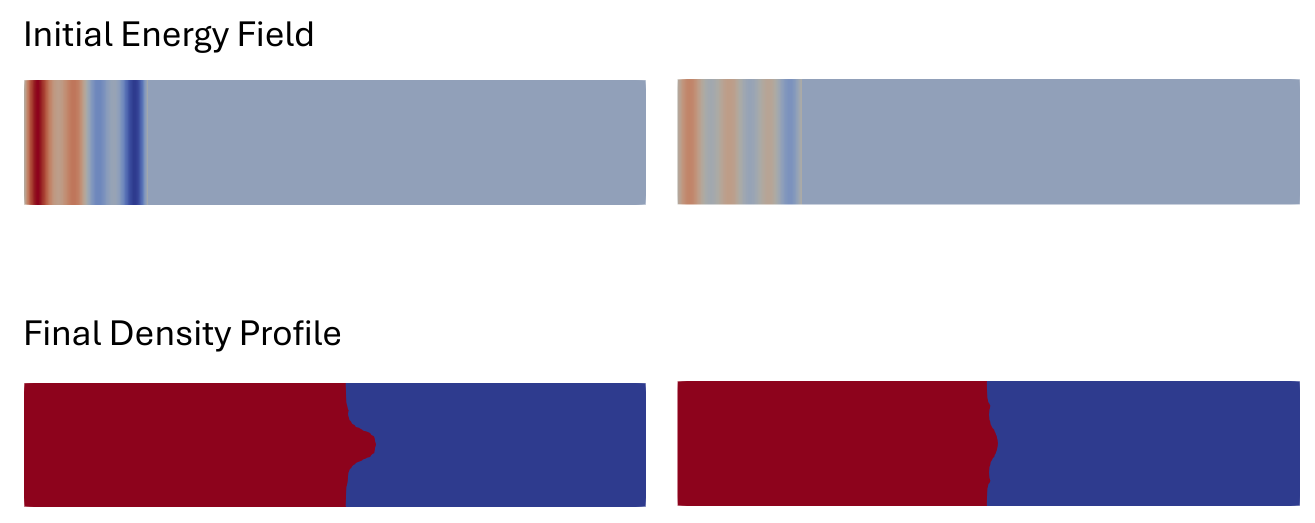}
    \caption{%
        Design exploration results for RMI suppression. Top row: initial energy
        field. Bottom row: final density profile. Left: configuration at the
        start of exploration (QoI = 4.1). Right: best design found by MADA
        (QoI = 3.7), showing significantly reduced jetting at the interface.
    }
    \label{fig:2d_rmi_laghos}
\end{figure}

For the problem of interest, we configure Laghos to simulate an RMI
using Mie-Gruneisen material models. In our setup, the computational domain consists of two materials, steel on the left and plastic on the right, 
separated by a sinusoidally perturbed interface. Within the steel, we designate a subregion as the design domain, 
where we control the initial energy distribution at runtime via parameters passed through the JMA. 
The prescribed energy profile in this design domain generates shocks that interact with the material interface 
and drive the development of the RMI. We treat all boundaries as sliding boundaries, which permit tangential motion 
along the boundary while preventing normal flow across it. We chose the setup shown in Figure~\ref{fig:2d_rmi_laghos} 
to study controllability of the interface evolution and, in particular, to investigate 
strategies for suppressing the growth of the RMI. Periodic diagnostic outputs from Laghos provide MADA 
with measurements of the instability length, enabling 
analysis of its dependence on the control parameters supplied by the JMA.
We use the Laghos3 simulation code (Section~\ref{sec:laghos}) to model shock propagation and RMI growth at the perturbed copper-plastic interface. 
The first region is initialized with a temperature profile characterized by a base internal energy of 0.1, to 
which sinusoidal perturbations along the x-axis are applied, subject to the constraint that local internal energy 
remains non-negative. The remaining domains are initialized with zero energy and velocity. 

\subsubsection{Design Parameters and Objectives}

The goal of the optimization is to identify initial energy field configurations that minimize RMI growth while
maximizing interface acceleration. Each design is parameterized by four values that define the sinusoidal 
energy initialization along the x-axis, characterizing the perturbation applied to the density interface. 
The quantity of interest guiding the optimization is formulated as 
$\text{QoI} = 0.5\lambda_1(x_2 - x_{\text{outer}})^2 + \lambda_2/(\delta + |v_{\text{ave}}|)$, 
where $x_2$ denotes the deformed x-coordinate at the interface mid-height, 
$x_{\text{outer}} = 0.5(x_1 + x_3)$ represents the average position of the top ($x_1$) 
and bottom ($x_3$) interface points (providing a symmetric reference), and 
$v_{\text{ave}} = (v_1 + v_2 + v_3)/3$ is the mean x-velocity across these three monitoring points. 
The first term ($\lambda_1 = 30.0$) penalizes lateral displacement of the interface midpoint from its 
symmetric reference position, thereby suppressing perturbation amplitude growth characteristic of RMI. 
The second term ($\lambda_2 = 4.0$) inversely rewards higher interface velocities with regularization 
parameter $\delta = 1.0$ preventing singularities, thereby promoting configurations that enhance 
shock-driven acceleration. 

\subsubsection{Workflow}

The user provides to MADA the problem description, the design parameters and their ranges via natural language.
MADA parses this input and orchestrates workflow between the various agents. 
All agents use OpenAI's o3 model for this study.
The GA generates a parameterized mesh
template through Cubit or PMesh. The IDA then produces a Latin hypercube
sampling plan for initial exploration (20 samples). The JMA
submits the ensemble to Tuolumne system via Flux and monitors job completion. 
As results return, the IDA extracts jet length from each run, ranks configurations, and identifies the
best-performing design.

For the second round, the IDA narrows the search to a
neighborhood around the current best and proposes additional samples. The cycle
repeats: sample generation, job submission, result collection, analysis. At the
end of each round, MADA reports a ranked table of configurations along with
visualizations of the top designs. The user can request further refinement or stop.
Over two exploration rounds, we evaluate 40 configurations.

% The best design reduced jet significantly as shown in Figure~\ref{fig:2d_rmi_laghos}.

\subsubsection{Results}

In Round 1, MADA evaluates 20 Latin hypercube samples spanning the design space. The IDA
analyzes the results and identifies configurations with lower QoI values, revealing that
certain combinations of sinusoidal parameters consistently produce lower RMI growth.
In Round 2, the IDA focuses sampling around the promising region from Round 1,
evaluating 20 additional configurations.

Figure~\ref{fig:2d_rmi_laghos} shows the optimization results. The initial configuration
yields a QoI of 4.1, while the best design MADA finds achieves a QoI of 3.7, representing
a 10\% improvement. The density field visualizations show reduced interface perturbation
amplitude in the optimized configuration compared to the initial design.

The entire optimization completes in approximately 40 minutes. Each Laghos
simulation runs for approximately 20 minutes, and each batch of 20 runs
executes in parallel, completing in about 20 minutes. With two rounds, the
total wall time is roughly 40 minutes. 
% LLM inference for analysis and planning adds approximately \$2 per round (\$4 total). 
The main savings come from human effort: without MADA,
a domain expert would need to manually configure each simulation, write job submission
scripts, monitor completion, parse output files, compute QoIs, and decide on sampling
strategies---a process that typically takes days for comparable design studies. MADA completes the same workflow in under an hour, accelerating the
design cycle and freeing domain experts to focus on interpreting results
and guiding the exploration.

% \begin{table}[t]
% \caption{Cost breakdown for the 40-configuration Laghos design study.}
% \label{tab:cost_tradeoff}
% \centering
% \footnotesize
% \begin{tabular}{lrr}
% \toprule
% Resource & MADA & Manual \\
% \midrule
% Wall time & 40 min & 2--3 days \\
% HPC allocation & 160 node-min & 160 node-min \\
% LLM inference & \$4 & --- \\
% Human effort & $<$10 min & 8--16 hours \\
% \bottomrule
% \end{tabular}
% \end{table}

% We also compare against a random sampling baseline. Uniform random sampling of 40
% configurations from the design space yields a best QoI of 3.85 on average (over 10 trials),
% compared to MADA's 3.7. The Latin hypercube initialization followed by targeted refinement
% outperforms pure random sampling by focusing evaluations on promising regions identified
% through the agent's analysis.
\label{sec:laghos-eval}

\subsection{Design Exploration with Surrogate Model for High Velocity Impact}
% \begin{figure}[t]
%     \centering
%     \includegraphics[width=0.8\linewidth]{figs/pchip_hvimpact.png}
%     \caption{%
%         Visualization of the high-velocity impact problem that results in the formation of RMI from~\cite{10.1063/5.0100100}.
%     }
%     \label{fig:hv_impact}
% \end{figure}

\begin{figure*}[t]
    \centering
    \begin{minipage}[t]{0.47\linewidth}
        \centering
        \begin{minipage}[c]{0.68\linewidth}
            \centering
            \includegraphics[width=\linewidth]{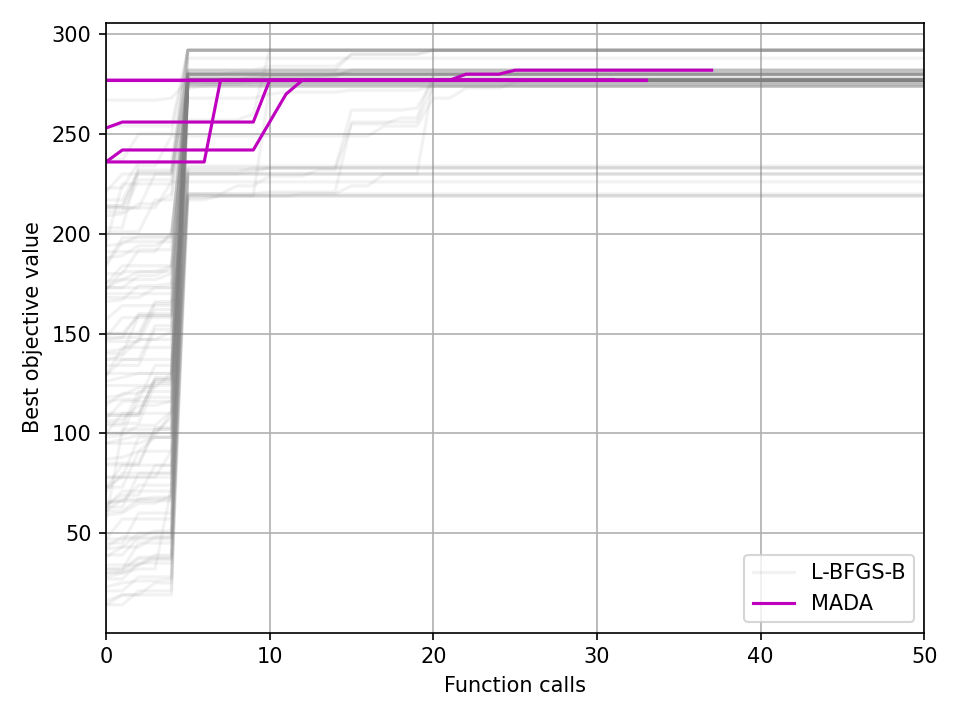}
        \end{minipage}
        \begin{minipage}[c]{0.30\linewidth}
            \centering
            \includegraphics[width=\linewidth]{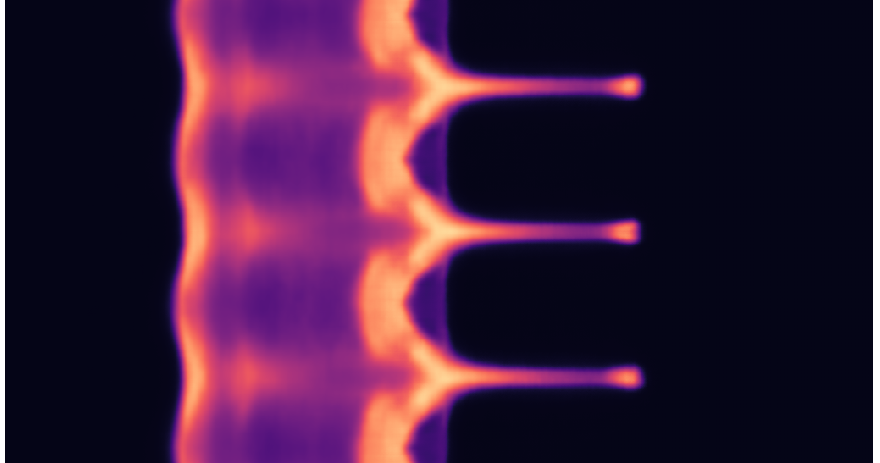}
        \end{minipage}
        \caption{%
            Maximizing RMI: comparison of 100 L-BFGS-B optimization runs against
            5 MADA runs. Each line shows the best objective versus number of
            evaluations (higher is better). MADA converges to a strong local
            optimum (objective 277). Right: density field for the best configuration found by MADA.
        }
        \label{fig:opt_results_max_rmi}
    \end{minipage}
    \hfill
    \begin{minipage}[t]{0.47\linewidth}
        \centering
        \begin{minipage}[c]{0.68\linewidth}
            \centering
            \includegraphics[width=\linewidth]{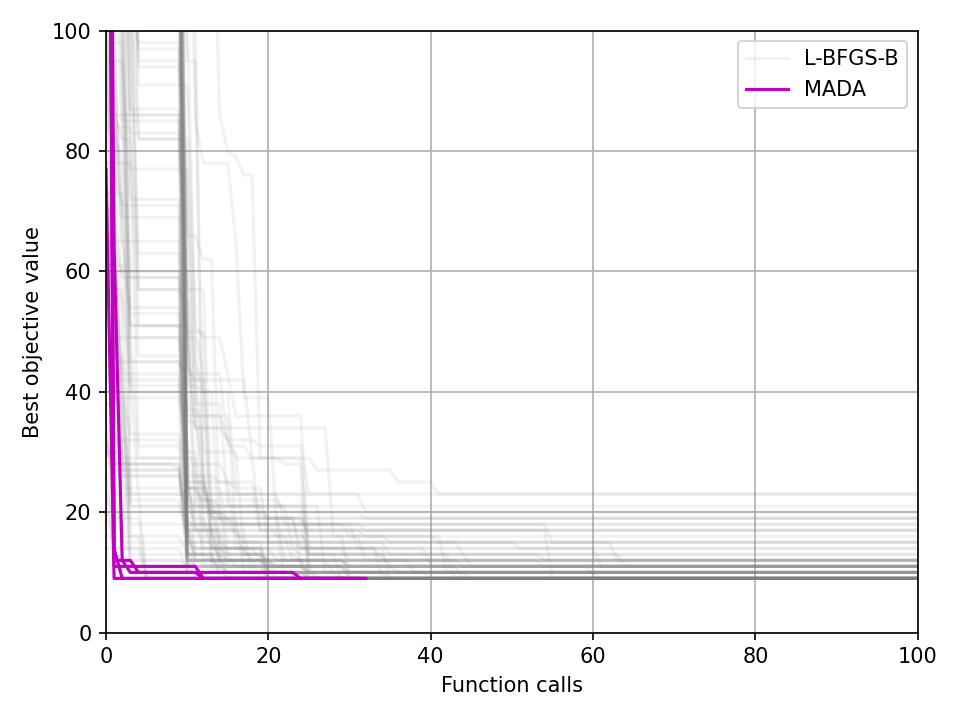}
        \end{minipage}
        \begin{minipage}[c]{0.30\linewidth}
            \centering
            \includegraphics[width=\linewidth]{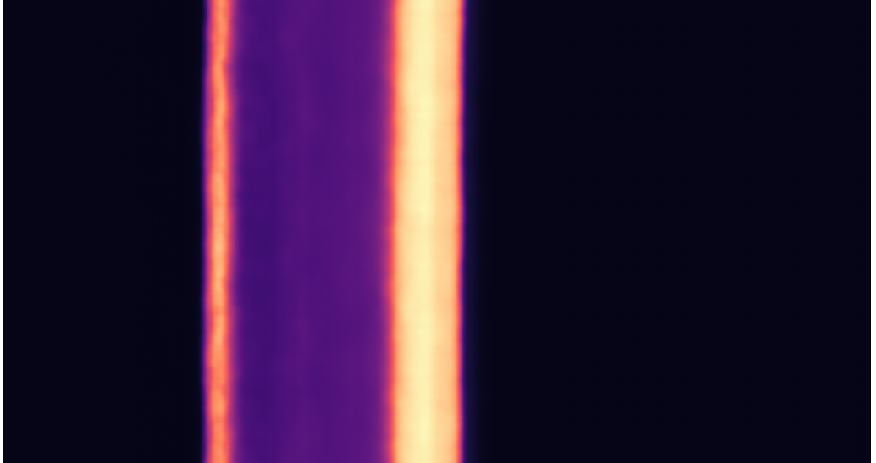}
        \end{minipage}
        \caption{%
            Minimizing RMI: comparison of 100 L-BFGS-B optimization runs against
            5 MADA runs. Each line shows the best objective versus number of
            evaluations (lower is better). MADA reaches the global optimum
            (objective 9) in under 40 evaluations across all runs.
            Right: density field for the best configuration found by MADA.
        }
        \label{fig:opt_results_min_rmi}
    \end{minipage}
\end{figure*}

% \begin{figure}[h]
%     \centering
%     \small
%     \begin{tabular}{llr}
%     \toprule
%     Round & Configuration $[P_1, P_2, P_3, P_4]$ & Objective \\
%     \midrule
%     1 & $[-0.22, +0.22, -0.22, +0.22]$ & 256 \\
%     1 & $[+0.22, -0.22, +0.22, -0.22]$ & 253 \\
%     1 & $[+0.25, +0.25, +0.25, +0.25]$ & 14 \\
%     1 & $[0.00, 0.00, 0.00, 0.00]$ & 11 \\
%     \midrule
%     2 & $[+0.25, -0.25, +0.25, -0.25]$ & \textbf{277} \\
%     2 & $[-0.25, +0.25, -0.25, +0.25]$ & \textbf{277} \\
%     2 & $[+0.22, -0.25, +0.25, -0.22]$ & 268 \\
%     \midrule
%     3 & $[+0.23, -0.24, +0.25, -0.25]$ & 274 \\
%     3 & $[+0.25, -0.24, +0.25, -0.24]$ & 273 \\
%     3 & $[+0.22, -0.25, +0.22, -0.25]$ & 264 \\
%     \bottomrule
%     \end{tabular}
%     \caption{Selected configurations from the agent's reasoning trace. Round 1
%     explores diverse patterns; Round 2 identifies the optimal alternating
%     configuration; Round 3 confirms the boundary optimum.}
%     \label{fig:agent_reasoning}
% \end{figure}

It is also possible to study RMI with the high velocity impact of two materials using
a surrogate model from~\cite{jekel2024machine} in place of a high fidelity multi-physics simulation code~\cite{rieben2020marbl}.
This approach enables rapid design
exploration without the computational cost of running full hydrodynamics
simulations, allowing MADA to evaluate hundreds of design
configurations in seconds rather than hours.

\subsubsection{Surrogate Model}

The surrogate is a neural network trained on 1,024 high-fidelity MARBL
simulations spanning the 4-dimensional design space~\cite{jekel2024machine}.
It takes four spline parameters as input and predicts the full 2D density
field, from which the RMI jet length objective is computed. Inference takes
$\sim$10 milliseconds per evaluation compared to $\sim$30 minutes for a full
simulation. The surrogate achieves a mean absolute error of $<$5\% on
held-out configurations, though it may extrapolate poorly outside the training distribution.
We recommend a two-stage workflow: use the surrogate for rapid
exploration to identify promising regions, then validate top candidates with full
simulations. This strategy combines the surrogate's speed advantage with the
simulation's physical fidelity. In our experiments, the surrogate correctly identified
the optimal region (alternating boundary values), which full simulations confirmed.

The underlying physics problem is a high velocity impact study that models
a Cu flier plate moving at 2~km/s and impacting a Cu target with perturbations at the free surface of the target.
As the shock wave reaches the interface perturbations, vorticity
deposition occurs along the interface due to misalignments between pressure and
density gradients. This creates an RMI that
results in jetting of the copper target material.
The inputs to the surrogate model are four spline points
that define the initial geometric perturbation on the free interface of the copper target. These spline points can be values in
$[-0.25, 0.25]$~cm on the free interface of the copper target using piecewise cubic Hermite interpolating polynomial. Initial geometric perturbations, caused by the spline points having different values, result in
RMI growth. In general, the larger the initial geometric perturbation, the larger the resulting RMI. Likewise, when all four
spline points are equivalent, there should be no initial geometric perturbation, resulting in little or no RMI growth.
% A depiction of this is shown in Figure~\ref{fig:hv_impact}.
Maximum RMI occurs when spline points alternate at extrema (e.g.,
$x=[-0.25, 0.25, -0.25, 0.25]$), creating the largest initial perturbation.
Minimum RMI occurs when all spline points are equal, producing no
perturbation. 
% We use the LLM agent instead of gradient-based optimization,
% as it provides interpretable reasoning about why configurations perform well.

\subsubsection{Workflow}

The user provides MADA with a natural language description of the problem: the
physical setup (high velocity impact of copper), the four spline parameters and their
ranges ($P_i \in [-0.25, 0.25]$~cm), and the optimization objective (minimize or maximize
RMI).  Instead of submitting jobs to HPC, the IDA queries
the surrogate model directly. The agent computes jet length
from the predicted density field by identifying the minimum
extent to which copper has penetrated into the air region.
All agents use OpenAI's o3 model for this study.

\subsubsection{Results}
MADA performs three rounds of refinement: an initial sample of configurations, followed by
two refinement rounds, for nearly 30 total evaluations with approximately 10 evaluations each round. Each round
completes in a few seconds. The IDA analyzes results after
each round, identifies promising regions, and proposes refined samples.

We compare 100 L-BFGS-B optimization runs from~\cite{jekel2024machine}
against 5 MADA runs. We ran 5 trajectories to assess reproducibility; since
all 5 converged consistently, additional runs were not necessary.
The results for minimizing RMI are shown in Figure~\ref{fig:opt_results_min_rmi} and the results for maximizing RMI are shown in Figure~\ref{fig:opt_results_max_rmi}.
Each L-BFGS-B line starts from a random point within the design space. 
The best observed RMI value is plotted against the total number of function evaluations. 

In all 5 runs when minimizing the RMI, the agent reaches the global optimum (objective 9) 
in less than 40 evaluations, matching the overall multi-start gradient-based optimization,
while providing interpretable reasoning.
The agent also appears to significantly outperform a single gradient-based optimization
by typically finding a better optimum in fewer function evaluations.
Maximizing RMI appears to be a much easier gradient-based optimization problem. The agent runs
tend to converge to a common optimum (objective 277) that many of the gradient-based optimizations find,
but they do not find the global optimum. This indicates that LLM-based
exploration is heuristic and could benefit from complementary optimization
techniques.

\subsubsection{Agent Reasoning Trace}
% basing this off of exp-flat-rmi-run4.txt
The agent's reasoning across three rounds reveals how it discovers the optimal
configuration through systematic exploration and analysis. % (Figure~\ref{fig:agent_reasoning}).

\paragraph{Round 1: Initial Exploration.}
The agent proposes 10 diverse configurations including alternating signs,
uniform values, and mixed patterns. After evaluation, it analyzes:

\begin{quote}
\small
\textit{``Large alternating signs dominate. The two best cases are strictly
alternating $\pm 0.22$, producing objective $\approx 250$---an order of magnitude
higher than uniform cases. Magnitude alone isn't enough: all-positive
$[+0.25, +0.25, +0.25, +0.25]$ and all-negative give very weak RMI ($\approx
11$--$14$). Therefore, merely big $P_i$ does not guarantee strong
mixing---sign variation is critical.''}
\end{quote}

\paragraph{Round 2: Focused Search.}
Based on Round 1 insights, the agent targets alternating patterns with large
magnitudes. After evaluation:

\begin{quote}
\small
\textit{``Magnitude matters: pushing $|P_i|$ to the upper bound (0.25) in an
alternating $\pm$ pattern produces the highest objectives ($\approx 270+$). The
best values sit on the boundary of the admissible range. 277 might already be
close to the global maximum.''}
\end{quote}

\paragraph{Round 3: Local Refinement.}
The agent tests micro-perturbations around the champion:

\begin{quote}
\small
\textit{``The global champion remains the fully-maximized alternating pattern
($|P_i| = 0.25$ with strict $\pm$ alternation). Micro-perturbations ($|P_i|
\approx 0.22$--$0.24$) sit consistently in 270--274 range---close but never
exceed 277. Objective appears to plateau near the boundary; perfect
symmetry/alignment at the extrema delivers the highest mixing.''}
\end{quote}

\noindent The agent's final recommendation for maximizing RMI: P = [+0.25, -0.25, +0.25, -0.25]
or [-0.25, +0.25, -0.25, +0.25].

This evaluation demonstrates that MADA can effectively explore design spaces
using surrogate models. The agent discovers that sign alternation
drives RMI growth. This capability to generate
explanations alongside solutions distinguishes MADA from traditional
optimization approaches.

\label{sec:professor-eval}

\subsection{Design Exploration with MARBL}
This case study exercises MADA's campaign steering capabilities on an RTI
problem simulated with MARBL~\cite{rieben2020marbl}, a production multi-physics
code. MADA manages a 1,800-run campaign over a mixed discrete-continuous
design space with 10 variables. Rather than following a predetermined
exploration plan, MADA analyzes intermediate results, proposes follow-on
studies focused on promising regions, and monitors execution.

\subsubsection{Problem Setup}

RTI occurs when a heavy fluid accelerates into a lighter fluid, causing the interface
to become unstable and develop characteristic mushroom-shaped structures.
The design space consists of 10 variables: eight continuous parameters
(\texttt{rho\_t}, \texttt{rho\_b}, \texttt{u0}, \texttt{drho}, \texttt{g}, \texttt{gamma},
\texttt{p0}, \texttt{x\_halfspan}) and two discrete mode variables (\texttt{kx\_mode}, \texttt{ky\_mode})
that control the initial perturbation state.

The objective is to minimize a mixing quantity of interest derived from the
volume fraction field $\alpha$, which measures the local proportion of heavy
fluid at each point ($\alpha = 1$ for pure heavy fluid, $\alpha = 0$ for pure
light fluid). The QoI is computed over the computational domain $\Omega$ with
area $A$:
\[
Q_{\mathrm{mix}} = \frac{1}{A}\int_{\Omega} 4\alpha(1-\alpha)\, dA,
\]
where lower values indicate less mixed (more separated) final states. This QoI
equals zero when fluids are fully separated and reaches its maximum when fluids
are completely mixed ($\alpha = 0.5$ everywhere).

\subsubsection{Adaptive Campaign Design}

The campaign began with a 200-run Latin hypercube sweep across all 10 variables.
MADA ranked the results and identified the two best designs:
run~0170 ($Q_{\mathrm{mix}} = 0.0244$) and run~0166 ($Q_{\mathrm{mix}} = 0.0272$).
These two designs used different discrete mode pairs, $(4,3)$ and $(6,2)$
respectively, which raised a natural question: would the good continuous
parameter values found near one design also work well with the other design's
discrete modes?

MADA proposed four follow-on studies. The two \emph{main} studies refine the
continuous parameters around each best design while keeping its original
discrete modes fixed. The two \emph{crosscheck} studies swap the discrete
modes between designs: they take one design's continuous parameter
neighborhood but pair it with the other design's discrete modes, testing
whether the good performance comes from the continuous values or the mode
choice.
Table~\ref{tab:rti_campaign} summarizes the four follow-on studies and their
outcomes.

% \begin{figure}[t]
%     \centering
%     \includegraphics[width=\linewidth]{figs/best2_worst2_density_final.png}
%     \caption{%
%         Final-time RTI density snapshots from the initial 200-run study.
%         Top row: best designs (0170 and 0166) showing separated fluid layers.
%         Bottom row: worst designs (0157 and 0018) showing significant mixing.
%         The mixing QoI penalizes the disordered states in the bottom row.
%     }
%     \label{fig:rti_density}
% \end{figure}

\subsubsection{Workflow and Monitoring}

MADA uses JMA to generate input files for each
study and to launch them on Flux. Rather than
launching all runs at once, MADA submitted the campaign in waves. After each
set, JMA uses \texttt{check\_job\_status} to summarize the state of the four
studies (pending, running, completed, and failed counts). Following the completions
of the jobs, the IDA analyze completed results.

The full campaign comprised approximately 1800 runs across initial and
follow-on studies, with individual MARBL simulations completing in 10--15
minutes. MADA handled the campaign design and analysis autonomously, with
approval checkpoints before launching each follow-on study. Agents used
OpenAI's GPT-5.4 model for this study.

% \begin{figure}[t]
%     \centering
%     \includegraphics[width=\linewidth]{figs/rti_all_runs_parameter_comparison.pdf}
%     \caption{%
%         Parameter comparison across the 1800-run RTI campaign. Stars indicate
%         the best designs; squares indicate the worst. The overall best result
%         ($Q_{\mathrm{mix}} = 0.0179$) came from the \texttt{anchor\_0166\_crosscheck}
%         branch, demonstrating that crosscheck branches can outperform main
%         refinement branches.
%     }
%     \label{fig:rti_params}
% \end{figure}

\subsubsection{Results}

Table~\ref{tab:rti_campaign} summarizes the campaign structure and outcomes.
The most significant finding is that the best observed design overall
($Q_{\mathrm{mix}} = 0.0179$) emerged from the \texttt{anchor\_0166\_crosscheck}
study, the smallest of the four follow-on studies with only 60 runs.

\begin{table}[t]
\caption{RTI campaign structure and best observed results per study.}
\label{tab:rti_campaign}
\centering
\scriptsize
\begin{tabularx}{\linewidth}{@{}l c c X@{}}
\toprule
Study & Runs & Best $Q_{\mathrm{mix}}$ & Purpose \\
\midrule
Initial & 200 & 0.0244 & Global 10-variable sweep \\
0170 main & 1000 & 0.0202 & Refine around 0170 with modes $(4,3)$ \\
0170 crosscheck & 120 & 0.0329 & 0170's neighborhood with 0166's modes $(6,2)$ \\
0166 main & 420 & 0.0204 & Refine around 0166 with modes $(6,2)$ \\
0166 crosscheck & 60 & \textbf{0.0179} & 0166's neighborhood with 0170's modes $(4,3)$ \\
\bottomrule
\end{tabularx}
\end{table}

This result validates MADA's crosscheck strategy: the best design was not a direct
refinement of the initial best anchor, but rather emerged from testing
discrete mode transfer between local neighborhoods. Without the crosscheck studies,
the campaign would have converged to a suboptimal region of the design space.

% Figure~\ref{fig:rti_density} shows representative density snapshots illustrating
% the difference between high-quality (separated) and low-quality (mixed) designs.
% Figure~\ref{fig:rti_params} visualizes how the best and worst designs cluster
% across the 10-dimensional parameter space, revealing that lower gravitational
% acceleration (\texttt{g}) and specific mode combinations correlate with
% reduced mixing.

This study shows that MADA can manage a large-scale campaign of 1,800 runs, 
tracking job status across all four follow-on
studies, and analyzing results as they complete. More importantly, MADA
reasoned about the structure of the design space, recognized that the
interaction between discrete modes and continuous parameters was uncertain,
and proposed the crosscheck studies to test it. That decision found the best
design overall. A standard refinement around the single best initial design
would have missed it.

\label{sec:marbl-eval}

\section{Observations and Lessons Learned}
\label{sec:observations}
The following distills what we observed across evaluations and what we learned
from building and running MADA with domain scientists on production HPC systems.

\noindent\textbf{End-to-end automation reduces human effort.} In the Laghos study, MADA
completed a 40-configuration design study in under an hour, and the MARBL campaign scaled 
this to 1,800 runs with only approval checkpoints between studies. In both cases, the user
provided a natural language problem description and MADA handled mesh
generation, job orchestration, and result analysis.

\noindent\textbf{Reasoning about results leads to better exploration strategies.}
Rather than following a fixed sampling plan, the IDA analyzes intermediate
results and adapts its strategy. All 5 surrogate runs reached the global
optimum for RMI minimization, while many of the 100 gradient-based runs
terminated at local minima. In the MARBL campaign, the IDA reasoned about the
interaction between discrete modes and continuous parameters, proposed the
crosscheck studies, and found the best overall design.

\noindent\textbf{Specialization through multi-agent decomposition.} Splitting the
workflow across JMA, GA, and IDA keeps each agent in a focused context,
avoiding the context window limits that would constrain a single monolithic
agent on a 1,800-run campaign. This decomposition also allows using different
models per agent: we used o3 for Laghos and surrogate studies, and GPT-5.4 for
the MARBL campaign.

\noindent\textbf{MCP server development is the primary integration cost.}
Exposing HPC tools to LLM agents via MCP servers is the main upfront effort.
Once developed, servers
are reusable: the same Flux MCP server managed job submission and monitoring
for both the Laghos and MARBL evaluations. 
Applications that accept configuration through input files (e.g., Laghos's Lua
decks) without requiring recompilation are particularly well-suited, as the MCP
server only needs to write new input files with the agent's design parameters. 
% Adding a new simulation code requires only a new MCP
% server, not modifying the core system or retraining models.

\noindent\textbf{Bounded autonomy is more useful than full autonomy.} Early experiments
with fully autonomous agents led to wasted compute on poorly conceived runs.
Adding approval checkpoints for major decisions, such as launching each
follow-on study in the MARBL campaign, struck a productive balance: agents
handle routine orchestration while experts retain control over costly choices.

\noindent\textbf{Narrow tool interfaces reduce agent errors.} Agents perform reliably
when MCP tools expose focused operations (e.g., \texttt{submit\_jobs\_async},
\texttt{get\_qoi}) rather than broad APIs. 
% The most common agent failure mode
% was not hallucination but proposing parameter values slightly outside valid
% ranges. 
Input validation in the MCP server layer catches cases where parameter values
are incorrect and returns structured errors, allowing the agent to correct and resubmit.

% \noindent\textbf{Flux's hierarchical scheduling suits agent-driven ensembles.} By
% obtaining a single large allocation and using a Flux sub-instance for ensemble
% management, we eliminated queue wait time between rounds. In a traditional
% batch workflow, each round would require separate submissions with potentially
% hours of queue delay. With Flux, round-to-round turnaround was under 2
% minutes, dominated by LLM reasoning rather than scheduling.

\noindent\textbf{Domain knowledge enters through prompts and tools, not training.} MADA
uses pre-trained LLMs directly without fine-tuning. Agent behavior is shaped by
system prompts, tool descriptions, and RAG-augmented context
(Section~\ref{sec:ga}). Encoding domain constraints (e.g., valid parameter
ranges, mesh quality requirements) directly into prompts proved more effective
than relying on the LLM's general knowledge.

% \noindent\textbf{LLM stochasticity is manageable with visibility.} As with any
% stochastic model, different runs may produce variations in sampling strategies.
% However, all 5 surrogate runs reached the global optimum for RMI minimization
% (Section~\ref{sec:professor-eval}). Reasoning traces provide full visibility
% into agent decisions, and the expert-in-the-loop interface enables verification
% at any workflow stage. A complete 40-configuration Laghos study costs
% approximately \$4--\$8 in LLM inference, making iteration affordable.

\noindent\textbf{Interpretable reasoning has practical value beyond optimization.} In
the surrogate study, the agent's analysis that ``sign variation is critical,
not just magnitude'' gave scientists a concise explanation of the design space
structure. In the MARBL campaign, the agent's rationale for proposing
crosscheck studies helped experts evaluate the strategy before approving it.
This transparency increased expert confidence.
%  and reduced the need forintervention.

\section{Related Work}
\label{sec:related}
Recent work has explored the application of LLMs to scientific computing and
discovery. ChemCrow~\cite{m2024augmenting} demonstrated how LLM-based agents
can automate chemical research by integrating computational chemistry tools,
while Boiko et al.~\cite{boiko2023autonomous} developed an autonomous
laboratory system where LLMs control physical experiments through robotic
platforms. In materials science, LLMs were applied for property prediction
and inverse design~\cite{jablonka2024leveraging,miret2024llms}. In the fusion context,
a multi-agent system was used to optimize fusion target design~\cite{shachar2025multi}.

Toward more general-purpose systems, several efforts have shown how LLM-based agents 
could support broader scientific and engineering
workflows. Google's AI Co-Scientist~\cite{gottweis2025towards}, a multi-agent system built on
Gemini 2.0, generated novel research hypotheses and proposals, reducing
hypothesis generation time from weeks to days and validating drug repurposing
candidates for acute myeloid leukemia. AlphaEvolve~\cite{novikovalphaevolve}
combined evolutionary search with LLMs to design advanced algorithms, discovering
improved matrix multiplication methods and optimizing Google's data center
scheduling. Similarly, Cheng et al.~\cite{cheng2025barbarians} showed that LLM-based
evolutionary search can match or exceed human-designed algorithms for systems
problems like scheduling and load balancing. More recently, URSA framework~\cite{grosskopf2025ursa} 
introduced a universal research and scientific agent that demonstrates the potential 
of AI-powered agents in automating scientific research workflows. These efforts demonstrate the
growing capability of LLM-based agents to accelerate scientific discovery
across diverse domains.

Multi-agent frameworks have gained traction for complex problem-solving.
The Microsoft Agent Framework (formerly AutoGen)~\cite{wu2024autogen} provides
a platform for building conversational multi-agent systems, which we extend
for HPC environments. CAMEL~\cite{li2023camel}
explores role-playing between agents for task completion, while
MetaGPT~\cite{hong2023metagpt} structures multi-agent collaboration like a
software company. Similarly, AgentVerse~\cite{chen2023agentverse} and
ChatDev~\cite{qian2023communicative} apply
multi-agent approaches for software development tasks.

In HPC contexts, workflow management systems like
Pegasus~\cite{deelman2015pegasus}, Merlin~\cite{PETERSON2022255}, 
Swift~\cite{wilde2011swift}, and
Fireworks~\cite{jain2015fireworks} have long automated computational
pipelines. However, these systems typically require explicit workflow
definitions and lack the adaptive capabilities of LLM-based approaches.

Our work differs from existing approaches by combining multi-agent LLM systems
with HPC workflow management. By leveraging MCP for tool integration, we enable
agents to interact directly with existing simulation codes, schedulers, and domain-specific tools.
In addition, we focus specifically on iterative design optimization workflows in scientific
computing. This integration enables dynamic adaptation, natural language
interaction, and intelligent orchestration of complex computational workflows
that would be difficult to achieve with traditional workflow systems or
single-agent approaches.

Notably, direct quantitative comparison with prior work is not possible because
no existing system combines these capabilities. Prior systems fall into two distinct
categories: (1) workflow managers (Pegasus, Swift, Fireworks, Merlin) that automate
job execution but require explicit workflow definitions and lack adaptive reasoning
based on intermediate results, and (2) LLM agents for scientific tasks (ChemCrow,
Co-scientist, URSA) that operate on individual tools or laboratory equipment but
do not integrate with HPC infrastructure, production simulation codes, or ensemble
job management. MADA is the first framework to unify LLM-based reasoning with
HPC job orchestration, mesh generation, and iterative design optimization in a
single system, making it a novel contribution without a direct baseline for comparison.

\section{Conclusion}
\label{sec:conclusion}
This paper presented the design, implementation, and evaluation of MADA on
Tuolumne, a leadership-class HPC system, for ICF-relevant hydrodynamic
instability problems. MADA integrates with production-grade HPC tools
including Flux for job scheduling, Cubit and PMesh for mesh generation,
Laghos and MARBL for simulation, automating the end-to-end orchestration of
meshing, job submission, monitoring, and result analysis that makes scientific
design workflows challenging. Across three evaluations, MADA drove iterative design with Laghos on HPC,
explored the design space using a surrogate model, and managed an 1,800-run
MARBL campaign where the best design emerged from an agent-proposed crosscheck
study.

As AI capabilities continue to advance, integrating LLM agents with HPC
infrastructure will become increasingly common for scientific discovery
workflows. Multi-agent decomposition, LLM reasoning over intermediate
results, and MCP-based tool integration provide a practical foundation for
building these systems, and we shared practical lessons from this deployment
to help practitioners leverage similar approaches. By integrating with existing HPC infrastructure and tools, MADA helps
scientists spend more time on scientific insight and less on orchestration,
accelerating scientific discovery.

\bibliographystyle{IEEEtran}
\bibliography{paper}

%%%%%%%%%%%%%%%%%%%%%%%%%%%%%%%%%%%%%%%%%%%%%%%%%%%%%
% AD/AE Appendix (does not count against 10-page limit)
%%%%%%%%%%%%%%%%%%%%%%%%%%%%%%%%%%%%%%%%%%%%%%%%%%%%%
% \newpage
% \input{ad_appendix}

\end{document}